# Semantically Redundant Training Data Removal and Deep Model Classification Performance: A Study with Chest X-rays

**Sivaramakrishnan Rajaraman[1], Ghada Zamzmi[1#], Feng Yang[1#], Zhaohui Liang[1], Zhiyun Xue[1], Sameer Antani[1*]**

[1]National Library of Medicine, National Institutes of Health, Bethesda, MD, United States

# Note: Dr. Ghada Zamzmi and Dr. Feng Yang contributed to this work while at the National Library of Medicine (NLM).

* Correspondence:
Sameer Antani
sameer.antani@nih.gov

**Keywords: chest X-rays, deep learning, semantic redundancy, informative sample selection, statistical significance.**

## Abstract

Deep learning (DL) has demonstrated its innate capacity to independently learn hierarchical features from complex and multi-dimensional data. A common understanding is that its performance scales up with the amount of training data. Another data attribute is the inherent variety. It follows, therefore, that semantic redundancy, which is the presence of similar or repetitive information, would tend to lower performance and limit generalizability to unseen data. In medical imaging data, semantic redundancy can occur due to the presence of multiple images that have highly similar presentations for the disease of interest. Further, the common use of augmentation methods to generate variety in DL training may be limiting performance when applied to semantically redundant data. We propose an entropy-based sample scoring approach to identify and remove semantically redundant training data. We demonstrate using the publicly available NIH chest X-ray dataset that the model trained on the resulting *informative* subset of training data significantly outperforms the model trained on the full training set, during both internal (recall: 0.7164 vs 0.6597, p<0.05) and external testing (recall: 0.3185 vs 0.2589, p<0.05). Our findings emphasize the importance of information-oriented training sample selection as opposed to the conventional practice of using all available training data.

## 1. Introduction

Deep learning (DL) models have demonstrated their superior capability by effectively extracting complex features and drawing meaningful inferences in medical image analysis tasks (1). For example, chest X-rays (CXRs) have been extensively studied for disease screening using DL. CXRs are among the most common imaging modalities used in medical practice, providing valuable insights into the diagnosis and monitoring of various conditions, such as pneumonia, COVID-19, lung cancer, tuberculosis, and other cardiopulmonary abnormalities (2,3). DL models are primarily used to analyze CXR images to solve unique and complex problems such as organ and disease region of interest (ROI) segmentation, detection, and classification tasks. The classification tasks can be binary (No Finding vs. Abnormal), multiclass (No Finding vs. Pneumonia vs. Tuberculosis), or multilabel (where a CXR may have multiple co-occurring abnormalities). In addition to image data, DL models can also analyze other types of data, such as radiological text reports, to improve classification performance (4).

Despite the promising performance of DL models applied to CXR analysis, there are challenges concerning the characteristics of data available for model training. The prevailing belief is that the performance of DL models scales up with the amount of training data. However, in medical computer vision, such large datasets are difficult to gather due to privacy and ethical concerns and the labor-intensive nature of manual annotation (2). Another significant challenge is the prevalence of semantic redundancy in medical image datasets. Semantic redundancy refers to the presence of visually similar or redundant information within the dataset. Such redundancies occur when multiple images containing nearly identical visual content or sharing similar characteristics, such as anatomical structures and/or disease manifestations, are present in the training data (5).

In medical image datasets, especially CXR images, semantic redundancies are common due to the standard practices in medical imaging. Multiple images of the same patient captured from slightly different perspectives or different moments in time may introduce semantic redundancy. For instance, the National Institutes of Health (NIH) CXR-14 dataset (3) could exhibit a considerable prevalence of semantic redundancy as the collection includes images captured from several patients who underwent multiple studies at the NIH Clinical Center (NIH-CC). Despite the impressive size of the dataset in terms of the overall number of films (n=112,120), while only a small percentage (7%) of patients in this data collection have more than 10 CXR films, these cases still constitute a substantial portion (45%) of the entire dataset. Consequently, the variability within the dataset might be significantly diminished, suggesting the possibility of a notable prevalence of semantic redundancy. Similarly, the MIMIC-CXR dataset (6) which is derived from the MIMIC-III critical care database, may also exhibit semantic redundancy due to multiple images taken at different time points during the patient's hospital stays. The presence of multiple images for the same patient needs evaluation for duplicative information which, in turn, may result in redundancy and, consequently, a learning bias and lead to model overfitting. In overfitting, the model focuses excessively on redundant patterns and memorizes these samples, while neglecting more diverse and informative examples, thus limiting its generalizability (7). The presence of semantic redundancy may also increase the computational burden during training, as the model processes similar information repeatedly, resulting in inefficiencies.

Researchers have proposed data augmentation methods (8,9) to improve model performance. Such augmentation techniques often rely on random affine transformations and may still generate images that are highly similar to existing samples which exacerbates the problem particularly when the dataset itself already exhibits a considerable degree of semantic redundancy. Steps to quantitatively assess the degree of semantic redundancy in the actual data must be taken before augmentation. Other approaches to reducing semantic redundancy involve using data sampling methods. Data sampling techniques proposed by (10) prioritize different training instances as learning progresses. The strategy, however, does not primarily aim at identifying and eliminating redundant data subsets but sampling the entire training set to accelerate the convergence rate. The method proposed by (11) divides the training set into numerous smaller subsets and iteratively removes these subsets until the classification performance dips below an acceptable level. The practicality of this method becomes questionable as it depends on iteratively creating numerous subsets from a single training set and evaluating performance that may increase the computational burden. The method proposed by (12) approaches the problem of subset selection as a constrained sub-modular maximization task.

However, (13) discovers that uncertainty-based sampling strategies outperform the method proposed by (12). The method proposed by (14) detects semantic redundancy by analyzing gradient magnitudes as a measure of importance. However, this method fails to differentiate between the near-duplicates existing in the training data if the resulting gradient magnitude is high. Another study (5) proposes a method to identify and remove redundant image samples through agglomerative clustering. However, the study presents significant challenges, including computational cost and the determination of the optimal number of clusters. Furthermore, the method does not inherently prioritize semantic informativeness, since the clusters may still contain redundant samples.

These studies are primarily conducted on general-purpose image datasets, not specifically tailored for medical images. Additionally, none of these studies address the issue of quantifying the *informativeness* of each training sample. This emphasizes the need for novel approaches in quantifying the informativeness of individual training samples within the context of medical image analysis. We hypothesize that a model trained on the subset of training data that is more informative than the entire training set may show reduced overfitting and improved performance. We propose to use an entropy-based scoring approach to quantify training sample informativeness. For this, we present a systematic method that involves: (i) training a DL model on the entire training set and assessing its baseline internal and external test performance; (ii) using entropy (15), which measures the uncertainty of the model's predictions, as a sample-wise scoring metric to assess each training sample's informativeness. Samples with high entropy are deemed informative, representing a case the model must learn more about, while those with low entropy are seen as less informative or redundant; (iii) selecting the informative training subset through a Bayesian optimization procedure and fine-tuning the model to achieve the least validation loss; (iv) comparing the performance of the model fine-tuned on this informative training subset, henceforth called the *Entropy* model, to the *Baseline* model trained on the entire training set, and (v) conducting statistical significance analyses to quantify the performance differences between these models. The Entropy model trained on the informative training subset is expected to exhibit improved performance and disease-consistent ROI localization.

The study is organized as follows: Section 2 discusses the materials and methods, Section 3 discusses the results, and Section 4 concludes the study and proposes future directions.

## 2. Materials and Methods

### 2.1. Datasets

We use two primary datasets in our retrospective analysis:

(i) National Institutes of Health (NIH) Chest X-ray-14 (NIH-CXR) dataset (3): This is a publicly available collection of de-identified CXR images, released by the NIH-CC. The dataset comprises 112,120 frontal-view CXR images collected from 30,805 distinct patients. Each CXR is paired with 14 labels indicative of cardiopulmonary diseases. The labels are derived from radiological reports via a Natural Language Processing (NLP) labeling tool. We use this as the internal dataset to train, validate, and test the DL model.

(ii) Shenzhen Tuberculosis (TB) CXR dataset (16): This publicly available anonymized dataset includes 326 CXRs showing normal lungs and 336 CXRs displaying a spectrum of TB-consistent manifestations. The CXRs are obtained during routine outpatient hospital procedures. The dataset is collected in collaboration with Shenzhen No.3 People's Hospital and Guangdong Medical College, Shenzhen, China. Each CXR is accompanied by a clinical reading that includes the patient's age, gender, and the details of lung abnormalities, if any. We use this dataset to perform external testing.

The NIH CXR dataset is divided at the patient level into 70% for training, 10% for validation, and 20% for testing to ensure the prevention of data leakage between different stages of model development. Table 1 shows the number of CXRs across the train, validation, and test sets.

**Table 1. Dataset characteristics.**

| Dataset | No Finding | | | Abnormal | | |
|---|---|---|---|---|---|---|
| | Train | Val | Test | Train | Val | Test |
| NIH-CXR (Internal) | 42288 | 6041 | 12083 | 36195 | 5170 | 10343 |
| Shenzhen-CXR (External) | -- | -- | 326 | -- | -- | 336 |

## 2.2. Lung delineation and cropping

The CXRs contain regions other than the lung ROI that do not offer any substantial value toward learning disease manifestations linked to the cardiopulmonary system. The presence of these irrelevant regions could inadvertently guide the model into learning unrelated feature representations. To tackle this issue, we employ a U-Net model with an Inception-V3 encoder backbone that we successfully implemented in one of our previous investigations (17) to segment the lung ROI. Subsequently, we delineate the lung boundaries using the predicted lung masks and crop them to fit within a bounding box that effectively contains all the pixels representing the lungs. We further resize the cropped CXRs to 224×224 pixel dimensions to match the input resolution that the VGG-16-based classification model (18) was originally trained on and normalize the image pixels within the range of [0, 1]. This process helps in handling variability in the pixel data and enhances the model's learning efficiency. The processing pipeline for UNet-based lung delineation and cropping is shown in Figure 1.

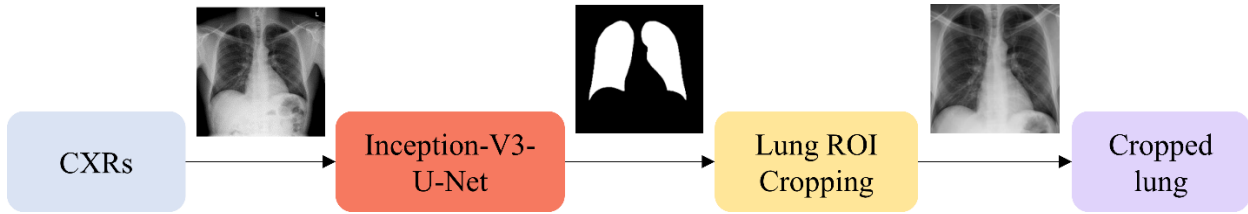

**Figure 1. UNet-based lung ROI delineation and cropping.**

## 2.3. Model architecture

We use the VGG-16 model architecture (18) due to its proficiency in medical image classification tasks, specifically CXRs (19–21). Rather than selecting the optimal DL model, we focus on proposing a sample-wise scoring method to identify the semantic redundancy in the training data.

# Removing semantic redundancy to improve deep model performance

The proposal can therefore be used with any DL model. We instantiate the VGG-16 model with its ImageNet-pretrained weights and truncate the model at its deepest convolutional layer. We add a global average pooling (GAP) layer to output the spatial average of the deepest convolutional feature maps following Equation (1).

$$GAP = \frac{1}{H \times W} \sum_{i=1}^{H} \sum_{j=1}^{W} f_{i,j} \qquad (1)$$

Here, $H$ and $W$ are the height and width of the input feature map, respectively, and $f_{i,j}$ is the value at the $i^{th}$ row and the $j^{th}$ column of the feature map. The output of the GAP layer feeds into the final dense layer with two nodes and Softmax activation to produce a probability distribution toward the task of classifying the CXRs as showing normal lungs or other cardiopulmonary-related disease manifestations.

We train the models in a synchronous distributed manner on four NVIDIA GeForce RTX 2080 GPUs, using a mini-batch size of 512. The training process unfolds in several stages. Initially, all convolutional weights are frozen while the final dense layer is randomly initialized and trained. We apply the Adam optimizer with an initial learning rate of $1\times10^{-3}$ to minimize the categorical cross-entropy loss, saving the model checkpoints when a decrease in validation loss is detected. Subsequently, we load the best weights from this first step, unfreeze the top convolutional block, and proceed to fine-tune it. The model now recompiled with the Adam optimizer at an initial learning rate of $1\times10^{-4}$, retains the same loss settings. The model checkpoint demonstrating the lowest validation loss is then utilized for internal and external testing. The test performance is evaluated at the threshold that optimizes the F-score on the validation dataset.

## 2.4. Informative Sample Selection Using Entropy-based Scoring

We use the entropy of the model's predictions to quantify the informativeness of each training sample. The entropy of a probability distribution is a measure of its uncertainty as given in Equation (2).

$$H(X) = -\sum P(X) \log_2(P(X)) \qquad (2)$$

Here, $X$ denotes a random variable, $P(X)$ denotes the probability mass function of $X$, and the summation is over all possible values of $X$. Applied to the probability distribution of a model's prediction for a specific input, entropy gives a measure of the model's prediction uncertainty. A model predicting all classes with equal probabilities has maximum entropy, indicating high uncertainty. Conversely, a model predicting a single class with full certainty has minimum entropy. High entropy samples are considered more informative, while low entropy samples may be redundant. The informative training subset is derived using a Bayesian optimization procedure.

## 2.5. Bayesian optimization for informative training subset selection

# Removing semantic redundancy to improve deep model performance

We apply Bayesian optimization to identify the optimal proportion of the informative training subset that achieves the least validation loss. This Bayesian approach to hyperparameter optimization, backed by the scientific literature (22,23), has proven its effectiveness in hyperparameter tuning. Unlike traditional grid and random searches that operate in discrete intervals, Bayesian optimization operates in a continuous interval, allowing for finer exploration of the search space. Our Bayesian optimization approach for identifying the optimal proportion of informative training subset operates as follows. First, we establish a search space in the continuous interval [0.5, 0.9], commencing the Bayesian optimization with 50 calls and 15 random starts. This selection ensures an efficient exploration without excessive resource consumption, with the optimal function value chosen based on minimal validation loss. For each function evaluation (i.e., the proportion of the training set), we undertake several steps, as illustrated in Figure 2.

```
Algorithm 1 Optimization and Training Algorithm
 1: Begin
 2: Step 1: Define continuous search space
 3: informative_sample_proportion in [0.5, 0.9]
 4: Step 2: Implement GP minimization
 5: Initialize Bayesian optimization with GP minimization
 6: Set total calls = 50
 7: Set random starts = 15
 8: for each call in total_calls do
 9:     if call ≤ random_starts then
10:         Randomly select values within the defined search space
11:     else
12:         Use Bayesian optimization to select next hyper-parameters based on prior observations
13:     end if
14:     if call == total_calls then break
15:     end if
16: end for
17: Step 4: Train subsequent model using selected samples
18: selected_samples = SelectinformativeSamples(Softmax_predictions)
19: Load pretrained model, freeze convolutional weights
20: Train last dense layer with randomly initialized weights using selected_samples
21: if validation loss increases then Store model checkpoint
22: end if
23: Step 5: Instantiate Step 4 model and fine-tune
24: model = InstantiateModel(best_weights_from_step_4)
25: Unfreeze last convolutional block of model
26: Train model with selected_samples
27: if validation loss < best validation loss so far then
28:     Store model checkpoint
29:     Store validation loss
30: end if
31: Step 6: Store optimized values and relevant information
32: Store(informative_sample_proportion)
33: Store(indices_of_most_informative_samples, corresponding_filenames)
34: End
```

Figure 2. Pseudocode for Bayesian optimization-based informative training subset selection.

We store the model checkpoints and record the value for the optimal proportion of the training subset, the training subset sample indices, and filenames at the end of the optimization process.

### 2.6. Performance and statistical significance evaluation

We assess model performance using key metrics: balanced accuracy, precision, recall, specificity, F-score, and Matthews Correlation Coefficient (MCC). These metrics provide crucial insights into the model performance across various medical vision tasks. Precision and recall reveal the model's capability to accurately identify true positive (TP) cases. Specificity measures the model's ability to predict true negative (TN) cases, i.e., when the model classifies the negative class samples correctly. The F-score unifies the precision and recall metrics into a single performance measure. The MCC metric provides a balanced measure for different class sizes, beneficial for imbalanced datasets often found in medical image recognition tasks.

We compare the statistical significance of recall achieved by the Baseline model which is trained with the full training set and the Entropy model which is trained on the informative training subset. We establish 95% binomial confidence intervals (CIs) using the Clopper-Pearson exact method and calculate $p$-values via the CI-based Z-test, a common technique used in statistical hypothesis testing. The standard error (SE) is computed as given in Equation (3).

$$SE = (CI_{upper} - CI_{lower})/(2 * 1.96). \tag{3}$$

Here, $CI_{upper}$ and $CI_{lower}$ represent the upper and lower bounds, respectively, of the 95% CIs for recall. The difference in recall ($\Delta Recall$) and SE ($\Delta SE$) between the compared models is determined as given in Equation (4) and Equation (5).

$$\Delta Recall = Recall2 - Recall1. \tag{4}$$

$$\Delta SE = sqrt(SE1^2 + SE2^2). \tag{5}$$

Here, *Recall1* and *Recall2* are the recall metrics, and *SE1*, and *SE2* are the SE metrics of the compared models. The Z-score is computed from these differences using Equation (6).

$$Z = \Delta Recall/\Delta SE. \tag{6}$$

The corresponding $p$-value for the Z-score is determined using an online Z-table calculator. A threshold of 0.05 establishes statistical significance. If the derived $p$-value is below this threshold, the difference is considered significant.

## 3. Results

### 3.1. Internal and external test performance

We showcase the effectiveness of the Bayesian approach in selecting the optimal proportion of the informative training subset via the results achieved through internal and external testing as shown in Table 2.

**Table 2. Internal and external test performance.** Here, *Baseline* rows show results from the model trained on the entire training data. The *Entropy* rows show the model trained on the optimal informative proportion (IP: 55.21%) of the training set identified through Bayesian optimization. The term *Bal. Acc.* denotes balanced accuracy. Bold numerical values denote superior performance in respective columns. The * identifies significantly superior performance for recall ($p<0.05$). The values in parenthesis denote 95% CI values for recall.

| Test | Model | Bal. Acc. | Precision | Recall | Specificity | F | MCC | IP | Z | p |
|---|---|---|---|---|---|---|---|---|---|---|
| Internal | Baseline | 0.7132 | **0.7077** | 0.6597 (0.6534,0.6666) | **0.7668** | 0.6829 | 0.4294 | | 12.875 | 0 |
| | Entropy | **0.7171** | 0.6848 | **0.7164*** (0.7105,0.7223) | 0.7178 | **0.7002** | **0.4332** | 0.5521 | | |
| External | Baseline | 0.6156 | **0.9062** | 0.2589 (0.2255,0.2923) | **0.9724** | 0.4027 | 0.3284 | | 2.3966 | 0.0165 |
| | Entropy | **0.6393** | 0.8917 | **0.3185*** (0.2830,0.3540) | 0.9601 | **0.4694** | **0.3615** | 0.5521 | | |

Our strategy using Bayesian optimization for detecting the informative proportion (IP), i.e., the most informative training subset that achieved the least validation loss, identified 55.21% (n=43,325) of the total training set (n=78,483, from Table 1).

We observe that with both the internal (NIH-CXR) and external (Shenzhen TB CXR) test sets, the Entropy model exhibits improved performance compared to the Baseline. Looking at the internal test performance, the Entropy model demonstrates a significantly superior recall (0.7164 vs 0.6597, $p<0.05$) and a relatively higher value for balanced accuracy, F-score, and MCC. A similar trend is observed with the external test using the Shenzhen TB CXR dataset. The Entropy model achieves significantly improved recall (0.3185 vs 0.2589, $p<0.05$) while maintaining a higher balanced accuracy, F-score, and MCC. This significantly improved recall metric implies that the Entropy model identifies a higher percentage of TP cases, a critical factor in any diagnostic tool. Although a slight decrease in precision and specificity is observed, superior values for recall, balanced accuracy, F-score, and MCC demonstrate improved internal and external performance using the Entropy model. These observations underscore the fact that just over half of the entire training set, when selected through our approach, contributes significantly to the model performance. The confusion matrices obtained with the internal and external test sets are shown in Figure 3.

**Removing semantic redundancy to improve deep model performance**

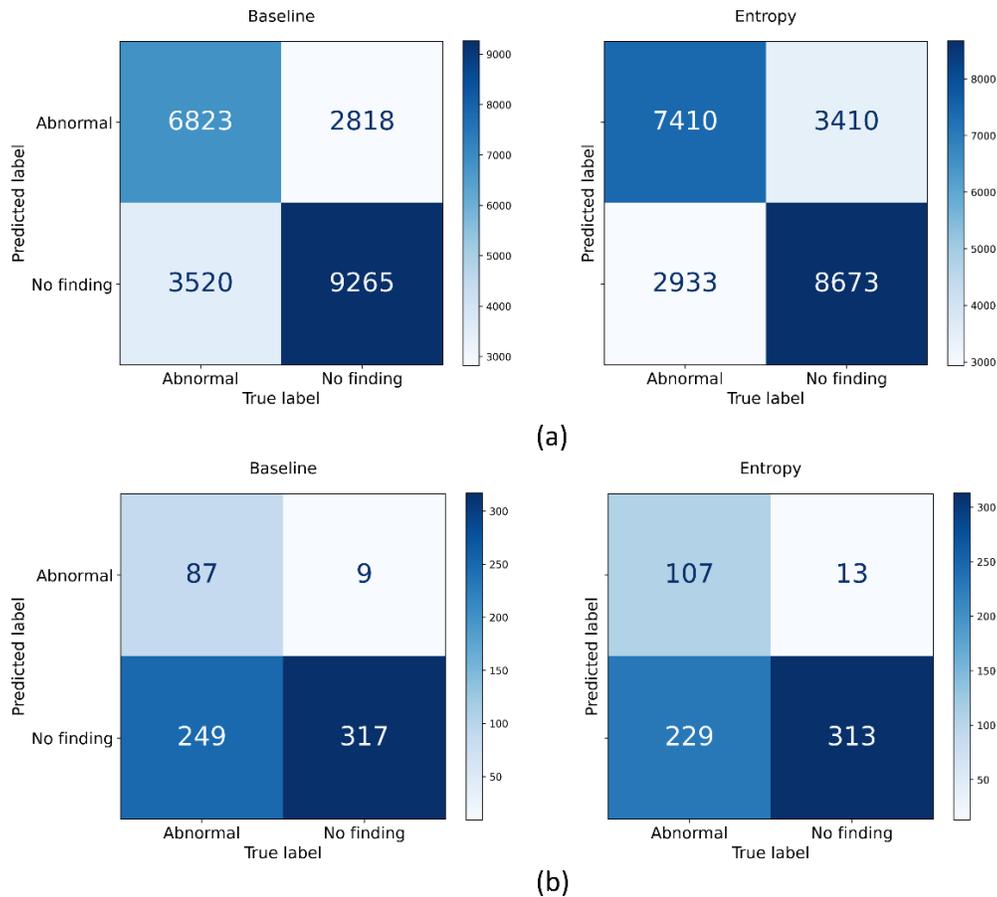

(a)

(b)

**Figure 3. Performance comparison between the Baseline and Entropy models.** (a) Confusion matrices for the internal test and (b) Confusion matrices for the external test.

Figure 4 shows a comparative examination of the training and validation loss curves for the Baseline and Entropy models. An ideal scenario involves a decreasing trend in both training and validation loss, signifying consistent learning and performance enhancement. In the Baseline model, we see a persistent decline in training loss. Interestingly, the validation loss initially drops until the 6$^{th}$ epoch, after which it begins to diverge from the training loss, indicating overfitting. The Entropy model, however, presents a more controlled learning process with a gradual decrease in training loss. Notably, the gap between the training and validation loss curves is smaller compared to the Baseline model, indicating less overfitting. This suggests that the Entropy model is superior at generalizing to the validation data, which aligns with the performance measures reported in Table 2.

### 3.2. Visualizing feature embeddings

We use t-distributed Stochastic Neighbor Embedding (t-SNE) to visualize the feature embeddings learned by the Baseline and Entropy models as shown in Figure 5. The t-SNE, a technique for dimensionality reduction, translates high-dimensional data into a two-dimensional space, enabling an insightful overview of feature distribution and class separation. These plots, generated with a perplexity of 40, a learning rate of 10, and 1000 iterations, reveal distinct differences between the Baseline and Entropy models. With both internal and external tests, the feature embeddings obtained from the Entropy model appear more compact and separated than those of the Baseline model. This suggests that the Entropy model more effectively discerns meaningful features for class differentiation, which is in line with its superior performance shown in Table 2.

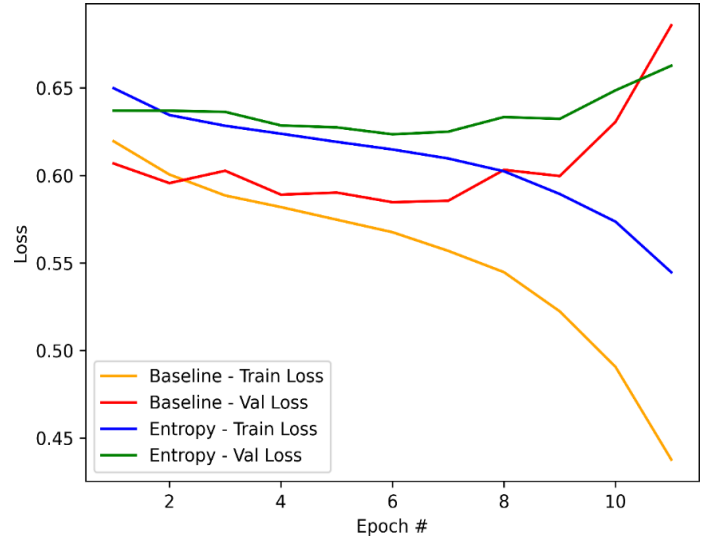

**Figure 4. Loss curves achieved with the Baseline and Entropy models.**

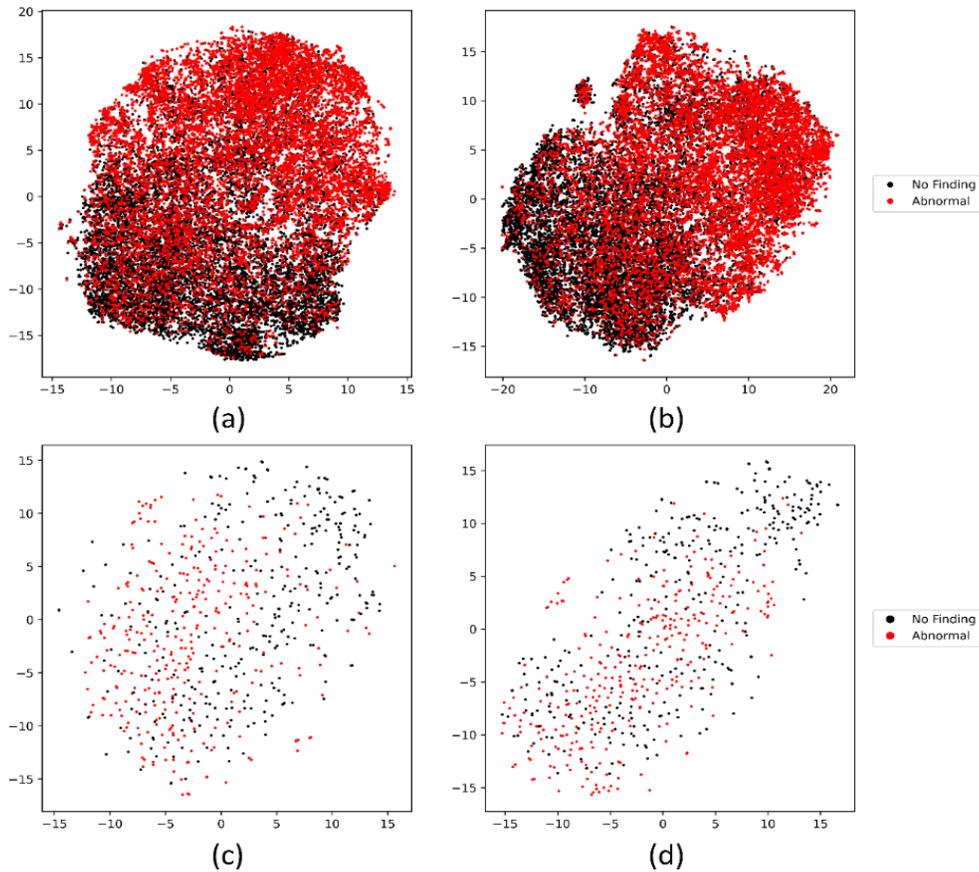

**Figure 5. The t-SNE embeddings from the baseline and entropy models.** (a) Baseline model with internal test; (b) Entropy model with internal test; (c) Baseline model with external test, and (d) Entropy model with external test.

## 3.3. Normalized Sankey flow diagrams

We use normalized Sankey flow diagrams to show a qualitative comparison of test performances by the Baseline and Entropy models (Figure 6). These diagrams effectively visualize both correct and incorrect predictions, facilitating performance comparison across models. The normalization technique assigns equivalent weights to classes on both the ground truth (left) and predictions (right) sides of the flow diagram, thereby ensuring equal representation for each class. The flow from one side of the diagram to the other, symbolized by the width of each strip, represents the model's prediction flow and the fraction of samples predicted to belong to a specific class. We observe that, with both internal and external tests, the Entropy model classifies a greater number of abnormal samples correctly. This is evident from the increased strip width corresponding to the abnormal class, and it aligns with the recall metrics presented in Table 2.

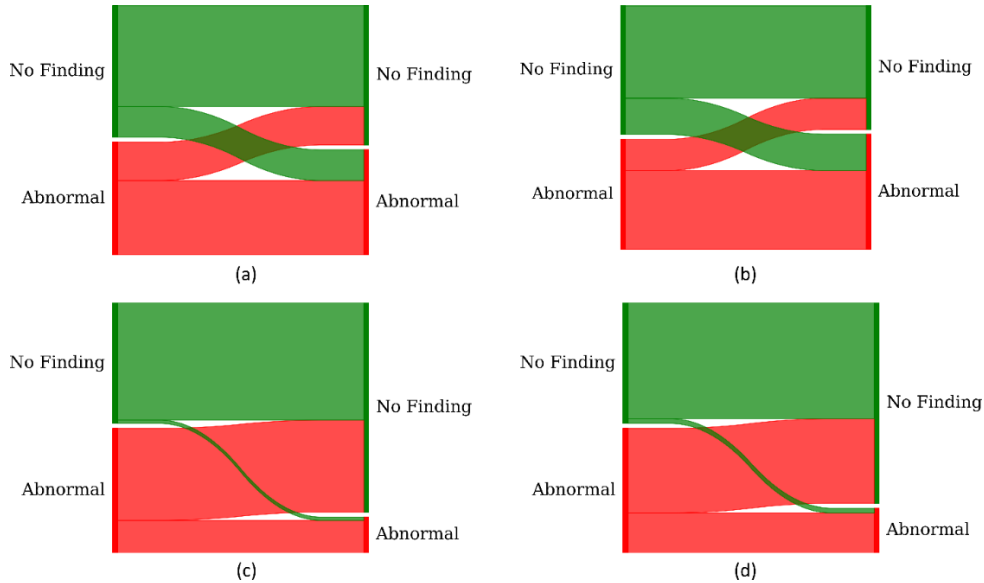

**Figure 6. Normalized Sankey flow diagrams.** The left side of the flow diagram denotes the ground truth, and the right side denotes the predictions. (a) and (b) Baseline and Entropy models predicting the internal test; (c) and (d) Baseline and Entropy models predicting the external test.

## 3.4. Histogram of entropy distribution

We present a histogram of entropy distributions for the informative training subset (comprising 55.21% of the samples) and the remaining or *redundant* samples in the training set (Figure 7). The x-axis represents entropy values (from 0 (minimal uncertainty) to 0.6931 (maximal uncertainty)), while the y-axis signifies the normalized count of samples corresponding to specific entropy values. The redundant and informative samples are depicted in green and red, respectively. The entropy values for the informative subset range approximately between 0.6917 and 0.6931, with a median value near 0.693 and a small standard deviation (around 0.000096). These metrics suggest that informative samples exhibit high entropy values, reflecting the model's uncertainty about these samples. Such samples are deemed valuable for learning as they may potentially encapsulate complex patterns or features that the model has yet to fully learn. Conversely, the redundant samples display a broad entropy range, indicating greater variability in entropy values. Lower entropy values suggest that the model is relatively confident about its predictions for these samples, implying that these samples might not offer novel information for learning. This depiction and associated analysis emphasize the argument for employing prediction entropy as a measure to identify informative samples and exclude redundant samples. Concentrating on informative samples that induce higher uncertainty may spur the model to learn more complex and varied features, potentially boosting its performance. We further calculate the mean entropy along with the 95% CIs. We compute the Z-score and the corresponding *p*-value to further ascertain the statistical significance of the apparent divergence in the entropy values between the informative and redundant training subsets. The *p*-value, being essentially zero (p=0.0000), underscores that the difference in entropy values between the informative and redundant subsets is statistically significant, suggesting a substantial shift in the mean entropy.

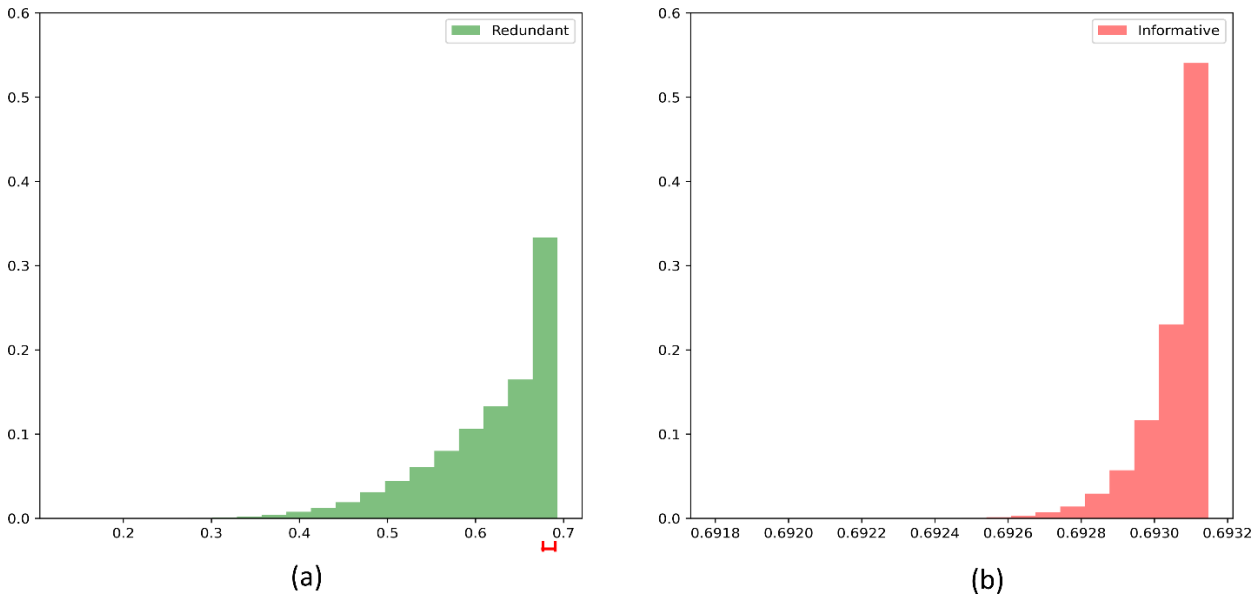

**Figure 7.** (a) Entropy distribution for the redundant training subset. The red bar shown on the x-axis denotes the range of entropy values measured for the informative training subset (0.6917 – 0.6931); (b) Entropy distribution for the informative training subset.

### 3.5. Class-selective relevance mapping (CRM)

We apply class-selective relevance mapping (CRM)-based visualization (24) to compare the learning behavior of the Baseline and Entropy models. CRM measures the impact of the deepest convolution layer's activations in localizing discriminative ROIs within the input image. Unlike traditional visualization tools like Grad-CAM (25), CRM calculates the linear sum of the incremental mean squared error (MSE) from all output layer nodes. The input sample's spatial elements with high CRM scores are identified as significant contributors to the classification task. Removing these elements would result in a substantial increase in the MSE at the output layer. The CRM-generated heatmaps overlaid on input images identify pixels learned as discriminative for classifying the input instances to their respective classes. Figure 8 displays CXR instances with overlaid CRM heatmaps, computed using the Baseline and Entropy models. Each CXR instance shows unique abnormal features indicative of cardiopulmonary disorders. The ground truth disease-consistent annotations are denoted by white bounding boxes. From these visualizations, it's apparent that compared to the Baseline model, the Entropy model aligns better with the ground truth boxes in localizing disease-consistent ROIs. This observation aligns with the Entropy model's improved recall performance, as shown in Table 2.

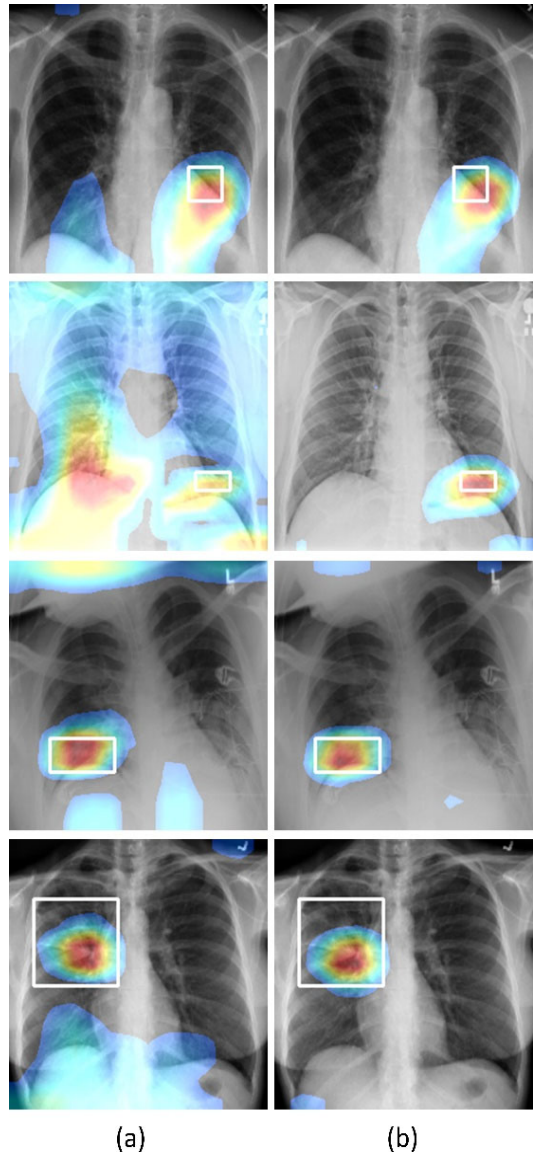

(a)        (b)

**Figure 8. CRM-based heatmaps overlaid on the CXR instances from the internal test.** (a) Baseline model and (b) Entropy model. The bounding boxes in white denote the ground truth disease-specific annotations.

### 3.6. Performance evaluation with the redundant training subset

Our study extends its evaluation to the models' performance on the redundant subset, the set of training samples not included in the informative training subset. This comparison sheds light on each

# Removing semantic redundancy to improve deep model performance

model's robustness and generalizability. Table 3 details the performance of the Baseline and Entropy models on this redundant subset. The Entropy model significantly outperforms the Baseline, notably in recall (0.7300 Vs. 0.6675, $p<0.05$), indicating its enhanced ability to correctly identify TP cases within the redundant subset. Despite a slight decrease in precision and specificity, higher balanced accuracy, F-score, and MCC highlight the Entropy model's superior performance.

**Table 3. Redundant test performance.** Bold numerical values denote significant performance in respective columns. The * denotes significantly superior performance for recall ($p<0.05$).

| Test | Model | Bal. Acc. | P | R | S | F | MCC | Z | p |
|---|---|---|---|---|---|---|---|---|---|
| Redundant | Baseline | 0.7260 | **0.7244** | 0.6675 (0.6625,0.6725) | **0.7845** | 0.6948 | 0.4559 | 17.8514 | 0 |
| | Entropy | **0.7352** | 0.7047 | **0.7300*** (0.7253,0.7347) | 0.7404 | **0.7171** | **0.4694** | | |

The comparison also extends to visual inspection of their learned feature embeddings using t-SNE. Figure 9 illustrates the t-SNE visualizations of the feature embeddings derived from the Baseline and Entropy models. Here, clear differences are observed: the Entropy model presents more compact, distinct feature clusters, whereas the Baseline model's feature space is less compact with a higher degree of class overlap, reflecting its comparatively inferior performance on the redundant subset.

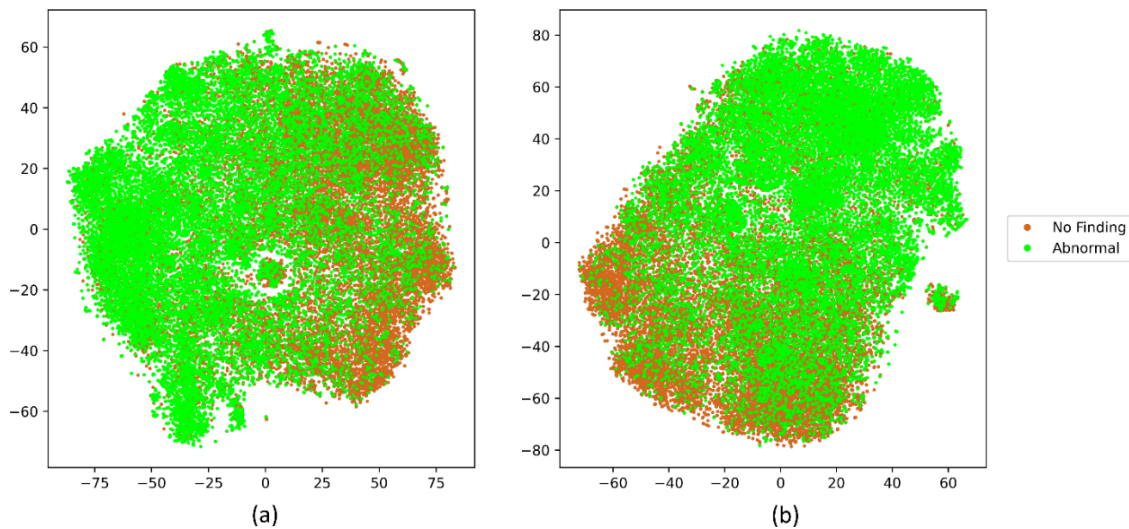

**Figure 9. The t-SNE embeddings are generated for the redundant subset.** (a) Baseline model and (b) Entropy model.

We further explore the models' prediction behavior using normalized Sankey flow diagrams, displayed in Figures 10(a) and 10(b) for the Baseline and Entropy models, respectively. Confusion matrices are presented in Figure 10(c). We note that the Entropy model accurately classifies a larger portion of abnormal samples, aligning with the recall metrics in Table 3.

**Removing semantic redundancy to improve deep model performance**

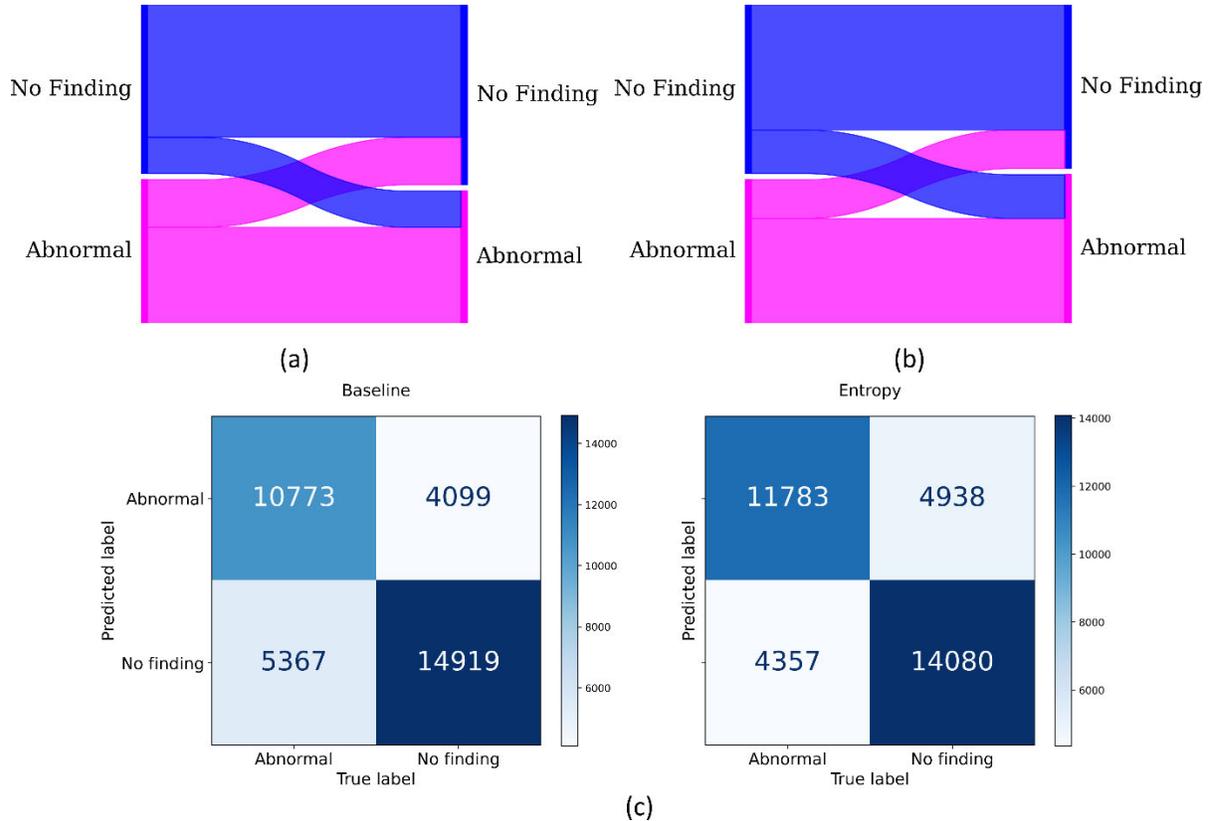

**Figure 10. Sankey flow diagrams for non-informative test performance.** (a) Baseline model; (b) Entropy model, and (c) Confusion matrices.

Using CRM visualizations, we analyze the learned behavior of both models in identifying disease-consistent ROIs in CXR instances from the redundant subset (Figure 11). The Entropy model demonstrates superior performance in aligning localization with ground truth bounding boxes, reinforcing its higher recall performance in Table 3, and underscoring its efficiency in identifying TP cases. In contrast, the Baseline model shows a decrease in localization performance, emphasizing its comparative deficiency in recall performance.

## 4. Conclusion and future work

Our approach combines entropy-based selection and performance-based optimization to select informative training samples, ensuring efficient training and reduced validation loss. Such training approaches would help discard less relevant or redundant data and lead to faster model development and deployment, accelerating the availability of AI-based diagnostic tools in healthcare. We observe that the Entropy model, trained on the informative training subset, surpassed the Baseline model, trained on the full training set, in both internal and external tests, and with the redundant set. The Entropy model showed significantly higher recall, which is vital for identifying TPs in medical diagnoses, meaning that more patients with the actual disease can accurately be detected. Improved generalization ensures that the model performs well on unseen data, making it more reliable in diverse clinical settings.

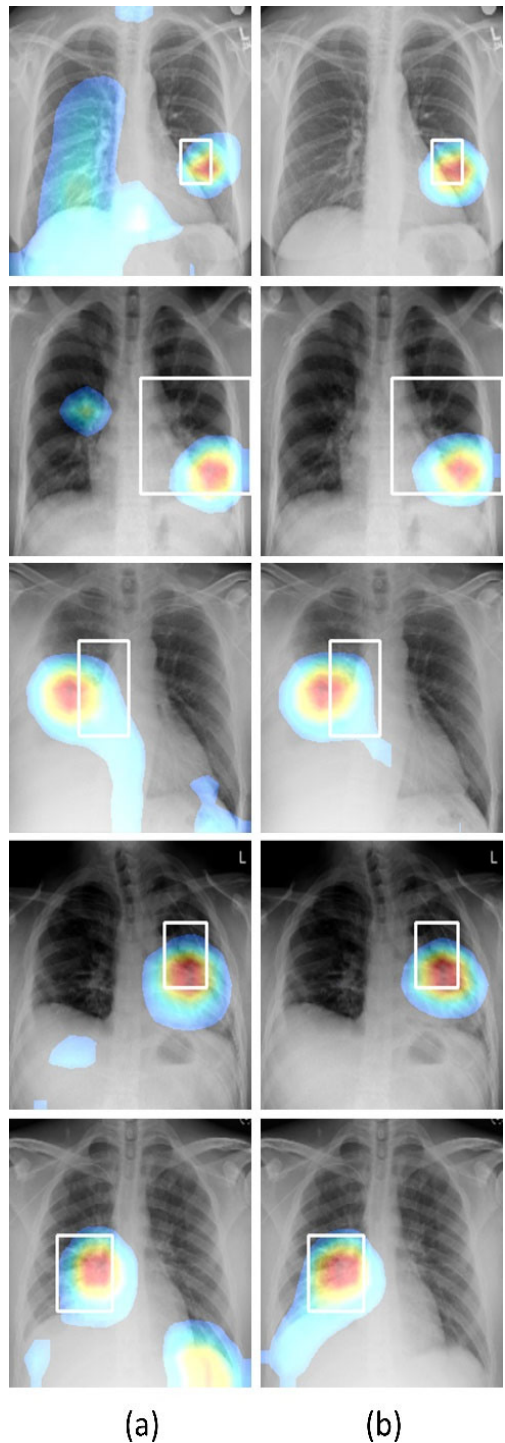

**Figure 11. CRM activations for sample CXR instances from the redundant subset.** (a) Baseline model and (b) Entropy model. The bounding boxes in white denote the ground truth disease-specific annotations.

It is indispensable to note that the entropy in the model's predictions does not always directly correlate with classification performance. In our study, the informative training subset is selected based on its impact on *validation performance*, not just entropy. Bayesian optimization considers the cumulative contribution to the model's validation performance. Rather than just selecting the highest entropy samples, it searches for a subset that, when used in training, results in the lowest validation loss. This may involve including high entropy samples and excluding others that don't contribute to reducing validation loss. Hence, while entropy aids in assessing sample informativeness, combining entropy-based selection and performance-based optimization ensures the selection of a truly informative training subset that improves model performance.

Our study diverges from clustering-based selection methods (5) by using an entropy-based sample-wise score to quantify the importance of each training sample. This approach, focusing on the most informative instances, is observed to enhance the model's classification performance and disease-consistent ROI localization. It also allows for more efficient training data utilization by omitting less relevant or redundant data.

Our study has several limitations. The focus is on binary disease classification, specifically distinguishing between CXRs as showing normal lungs and other cardiopulmonary abnormalities. The applicability of the entropy-based sample selection in varied medical imaging tasks and specific diseases remains to be proven. Furthermore, our approach doesn't consider label noise or other uncertainties which could affect model performance. Particularly in medical computer vision tasks, labels extracted from radiological reports using NLP tools can sometimes be noisy or incorrect, which can negatively impact the model, even when trained with high-entropy samples (3). Also, while entropy measures the model's prediction uncertainty, it doesn't differentiate between inherent data ambiguities and model limitations. High entropy might also be assigned to samples with visually diverse but diagnostically irrelevant features.

Future work shall consider additional scoring methods and optimization strategies that account for label quality, different types of uncertainties, and the relevance of data variations, along with entropy-based scoring. Future research shall focus on efficient, alternate strategies for selecting informative samples for model training. This would create a more robust training set, improving model performance and generalization. Integrating the entropy-based scoring approach with techniques like data augmentation or self-supervised learning could further enhance model performance. Despite these limitations, our study marks a significant step in harnessing an entropy-based sample scoring approach to improve model performance.

## 5. Conflict of Interest

The authors declare that the research was conducted in the absence of any commercial or financial relationships that could be construed as a potential conflict of interest.

## 6. Author Contributions

Sivaramakrishnan Rajaraman: Conceptualization, Data curation, Formal analysis, Methodology, Software, Validation, Visualization, Writing – original draft, Writing – review & editing; Ghada Zamzmi: Methodology, Writing – review & editing; Feng Yang: Methodology, Writing – review & editing; Zhaohui Liang: Methodology, Writing – review & editing; Zhiyun Xue: Conceptualization, Methodology, Writing – review & editing; Sameer Antani: Conceptualization, Formal analysis, Funding acquisition, Investigation, Project administration, Resources, Supervision, Validation, Writing – review & editing.

7.  **Funding**

This study is supported by the Intramural Research Program (IRP) of the National Library of Medicine (NLM) and the National Institutes of Health (NIH).

8.  **Data Availability Statement**

The datasets used in this study are publicly available and are cited within the manuscript text.

**References**

1. Lecun YA, Yoshua B, Geoffrey H, Bengio Y, Hinton GE. Deep learning. *Nature* (2015) **521**:436–444. doi:10.1038/nature14539

2. Irvin J, Rajpurkar P, Ko M, Yu Y, Ciurea-Ilcus S, Chute C, Marklund H, Haghgoo B, Ball R, Shpanskaya K, et al. CheXpert: A large chest radiograph dataset with uncertainty labels and expert comparison. *33rd AAAI Conf Artif Intell AAAI 2019, 31st Innov Appl Artif Intell Conf IAAI 2019 9th AAAI Symp Educ Adv Artif Intell EAAI 2019* (2019)590–597. doi:10.1609/aaai.v33i01.3301590

3. Wang X, Peng Y, Lu L, Lu Z, Bagheri M, Summers RM. ChestX-ray8: Hospital-scale Chest X-ray Database and Benchmarks on Weakly-Supervised Classification and Localization of Common Thorax Diseases. in *The IEEE Conference on Computer Vision and Pattern Recognition (CVPR)*, 1–19. doi:10.1109/CVPR.2017.369

4. Harmon SA, Sanford TH, Xu S, Turkbey EB, Roth H, Xu Z, Yang D, Myronenko A, Anderson V, Amalou A, et al. Artificial intelligence for the detection of COVID-19 pneumonia on chest CT using multinational datasets. *Nat Commun* (2020) **11**:4080. doi:10.1038/s41467-020-17971-2

5. Birodkar V, Mobahi H, Bengio S. Semantic Redundancies in Image-Classification Datasets: The 10% You Don't Need. (2019)

6. Johnson AEW, Pollard TJ, Berkowitz SJ, Greenbaum NR, Lungren MP, Deng C, Mark RG, Horng S. MIMIC-CXR, a de-identified publicly available database of chest radiographs with free-text reports. *Sci Data* (2019) **6**:317. doi:10.1038/s41597-019-0322-0

7. Shorten C, Khoshgoftaar TM. A survey on Image Data Augmentation for Deep Learning. *J Big Data* (2019) **6**:60. doi:10.1186/s40537-019-0197-0

8. Karki M, Kantipudi K, Yang F, Yu H, Wang YXJ, Yaniv Z, Jaeger S. Generalization Challenges in Drug-Resistant Tuberculosis Detection from Chest X-rays. *Diagnostics* (2022) **12**:1–23. doi:10.3390/diagnostics12010188

9. Ganesan P, Rajaraman S, Long R, Ghoraani B, Antani S. Assessment of Data Augmentation Strategies Toward Performance Improvement of Abnormality Classification in Chest Radiographs. in *Proceedings of the Annual International Conference of the IEEE Engineering in Medicine and Biology Society, EMBS* doi:10.1109/EMBC.2019.8857516

10. Katharopoulos A, Fleuret F. Not All Samples Are Created Equal: Deep Learning with Importance Sampling. in *Proceedings of the 35th International Conference on Machine Learning, {ICML} 2018, Stockholmsmässan, Stockholm, Sweden, July 10-15, 2018* Proceedings of Machine Learning Research., eds. J. G. Dy, A. Krause (PMLR), 2530–2539. Available at: http://proceedings.mlr.press/v80/katharopoulos18a.html

11. Ohno-Machado L, Fraser HSF, Øhrn A. Improving machine learning performance by removing redundant cases in medical data sets. in *{AMIA} 1998, American Medical Informatics Association Annual Symposium, Lake Buena Vista, FL, USA, November 7-11, 1998* (AMIA). Available at: https://knowledge.amia.org/amia-55142-a1998a-1.588514/t-001-1.590475/f-001-1.590476/a-099-1.590738/a-100-1.590735

12. Wei K, Iyer RK, Bilmes JA. Submodularity in Data Subset Selection and Active Learning. in *Proceedings of the 32nd International Conference on Machine Learning, {ICML} 2015, Lille, France, 6-11 July 2015* {JMLR} Workshop and Conference Proceedings., eds. F. R. Bach, D. M. Blei (JMLR.org), 1954–1963. Available at: http://proceedings.mlr.press/v37/wei15.html

13. Kaushal V, Sahoo A, Doctor K, Uppalapati NR, Shetty S, Singh P, Iyer RK, Ramakrishnan G. Learning From Less Data: Diversified Subset Selection and Active Learning in Image Classification Tasks. *CoRR* (2018) **abs/1805.1**: Available at: http://arxiv.org/abs/1805.11191

14. Vodrahalli K, Li K, Malik J. Are All Training Examples Created Equal? An Empirical Study. *CoRR* (2018) **abs/1811.1**: Available at: http://arxiv.org/abs/1811.12569

15. Van Erven T, Harrëmos P. Rényi divergence and kullback-leibler divergence. *IEEE Trans Inf Theory* (2014) doi:10.1109/TIT.2014.2320500

16. Jaeger S, Candemir S, Antani S, Wang Y-XJ, Lu P-X, Thoma G. Two public chest X-ray datasets for computer-aided screening of pulmonary diseases. *Quant Imaging Med Surg* (2014) **4**:475–477. doi:10.3978/j.issn.2223-4292.2014.11.20

17. Rajaraman S, Yang F, Zamzmi G, Xue Z, Antani S. Assessing the Impact of Image Resolution on Deep Learning for TB Lesion Segmentation on Frontal Chest X-rays. *Diagnostics* (2023) **13**: doi:10.3390/diagnostics13040747

18. Simonyan K, Zisserman A. Very deep convolutional networks for large-scale image recognition. in *3rd International Conference on Learning Representations, ICLR 2015 - Conference Track Proceedings*

19. Khan MA, Rajinikanth V, Satapathy SC, Taniar D, Mohanty JR, Tariq U, Damaševičius R. VGG19 network assisted joint segmentation and classification of lung nodules in CT images. *Diagnostics* (2021) **11**:1–16. doi:10.3390/diagnostics11122208

20. Rajaraman S, Zamzmi G, Folio L, Alderson P, Antani S. Chest x-ray bone suppression for improving classification of tuberculosis-consistent findings. *Diagnostics* (2021) **11**:1–21. doi:10.3390/diagnostics11050840

21. Ren S, He K, Girshick R, Sun J. Faster R-CNN: Towards Real-Time Object Detection with Region Proposal Networks. *IEEE Trans Pattern Anal Mach Intell* (2017) **39**:1137–1149. doi:10.1109/TPAMI.2016.2577031

22. Bergstra J, Bengio Y. Random Search for Hyper-Parameter Optimization. *J Mach Learn Res* (2012) **13**:281–305. doi:10.1162/153244303322533223

23. Snoek J, Rippel O, Adams RP. Scalable Bayesian Optimization Using Deep Neural Networks.

*Int Conf Mach Learn* (2015)

24. Kim I, Rajaraman S, Antani S. Visual interpretation of convolutional neural network predictions in classifying medical image modalities. *Diagnostics* (2019) **9**: doi:10.3390/diagnostics9020038

25. Selvaraju RR, Cogswell M, Das A, Vedantam R, Parikh D, Batra D. Grad-CAM: Visual Explanations from Deep Networks via Gradient-Based Localization. *Proc IEEE Int Conf Comput Vis* (2017) **2017-Octob**:618–626. doi:10.1109/ICCV.2017.74